\documentclass{article}

\PassOptionsToPackage{numbers}{natbib}

\usepackage[preprint]{neurips_2025}

\usepackage[utf8]{inputenc} %
\usepackage[T1]{fontenc}    %
\usepackage{hyperref}       %
\usepackage{url}            %
\usepackage{booktabs}       %
\usepackage{amsfonts}       %
\usepackage{nicefrac}       %
\usepackage{microtype}      %
\usepackage{xcolor}         %
\usepackage{amsmath}
\usepackage{multirow} 
\usepackage{booktabs} 
\usepackage{adjustbox}
\usepackage{amssymb}%
\usepackage{pifont}%
\usepackage{cleveref}
\usepackage{makecell}
\usepackage{graphicx}
\usepackage{subcaption}

\definecolor{mygreen}{RGB}{0, 150, 0}

\usepackage[draft,inline,nomargin,marginclue]{fixme}
\fxusetheme{colorsig}
\fxsetup{theme=color,mode=multiuser}
\FXRegisterAuthor{ds}{envds}{\color{orange}ds}
\FXRegisterAuthor{db}{envdb}{\color{blue}db}
\FXRegisterAuthor{gb}{envgb}{\color{violet}gb}

\title{DRAGON: A Large-Scale Dataset of Realistic Images Generated by Diffusion Models}

\author{%
    Giulia Bertazzini \\ Department of Information Engineering \\ University of Florence \\ \texttt{giulia.bertazzini@unifi.it} \\ 
    \And Daniele Baracchi \\ Department of Information Engineering \\ University of Florence \\ \texttt{daniele.baracchi@unifi.it} \\ \And Dasara Shullani \\ Department of Information Engineering \\ University of Florence \\ \texttt{dasara.shullani@unifi.it} \\ \And Isao Echizen \\ National Institute of Informatics \\ Tokyo, Japan \\ \texttt{iechizen@nii.ac.jp} \And Alessandro Piva \\ Department of Information Engineering \\ University of Florence \\ \texttt{alessandro.piva@unifi.it} \\\\
}

\begin{document}

\maketitle

\begin{abstract}

The remarkable ease of use of diffusion models for image generation has led to a proliferation of synthetic content online. While these models are often employed for legitimate purposes, they are also used to generate fake images that support misinformation and hate speech. Consequently, it is crucial to develop robust tools capable of detecting whether an image has been generated by such models. Many current detection methods, however, require large volumes of sample images for training. Unfortunately, due to the rapid evolution of the field, existing datasets often cover only a limited range of models and quickly become outdated. In this work, we introduce DRAGON, a comprehensive dataset comprising images from 25 diffusion models, spanning both recent advancements and older, well-established architectures. The dataset contains a broad variety of images representing diverse subjects. To enhance image realism, we propose a simple yet effective pipeline that leverages a large language model to expand input prompts, thereby generating more diverse and higher-quality outputs, as evidenced by improvements in standard quality metrics. The dataset is provided in multiple sizes (ranging from extra-small to extra-large) to accomodate different research scenarios. DRAGON is designed to support the forensic community in developing and evaluating detection and attribution techniques for synthetic content. Additionally, the dataset is accompanied by a dedicated test set, intended to serve as a benchmark for assessing the performance of newly developed methods.
\end{abstract}

\section{Introduction}

Machine learning technologies for image generation have exploded in popularity in recent years. Whereas generative models were once considered tools reserved for specialists, text-conditioned diffusion models are now accessible to virtually anyone. In addition to commercial services offered via APIs, open-weight generative models can be easily downloaded and run by users on their personal computers. While this democratization of access has facilitated legitimate uses, it has simultaneously lowered the barrier for malicious actors seeking to spread misleading content for propaganda or disinformation. Mitigating this phenomenon is particularly challenging when adversaries have full access to both the model's code and weights, and the ability to modify them. In recent instances, AI models have been exploited to generate false images of Donald Trump, Emmanuel Macron, and Julian Assange~\cite{sky-news}, raising serious concerns about the reliability of online visual content \cite{ai-meme}.

To address this challenge, researchers in multimedia forensics have developed techniques for detecting whether an image is synthetically generated and, when possible, attributing it to a specific generative model. However, many earlier methods were designed for images produced by generative adversarial networks and perform poorly on diffusion-generated images, as the subtle traces they rely on differ substantially between the two model families~\cite{corvi2023detection}.
Consequently, new methodologies have been required to address the challenges of detection and attribution in this evolving landscape, and as consequence there is a growing need to collect large synthetic image datasets to effectively train these models.
Existing datasets are often limited to a few generative models, reducing both the generalizability of detection methods and the effectiveness of attribution approaches.
Existing datasets quickly become outdated as generative techniques evolve, being limited to the methods available at their time of release. Moreover, many lack consideration for image realism. Although detection methods may not rely on semantic content, training on realistic images better reflects real-world conditions. In contrast, datasets of low-quality, trivially identifiable fakes offer limited value as benchmarks and risk encouraging models to rely on artifacts that may disappear as generation methods improve.

To overcome these limitations, we introduce DRAGON: a large-scale Dataset of Realistic imAges Generated by diffusiON models. To the best of our knowledge, DRAGON is the largest and most diverse dataset proposed to date for diffusion model detection and attribution tasks, comprising 2,600,000 synthetic images generated using 25 distinct diffusion models. The synthetic images were generated based on the 1,000 classes of ImageNet~\cite{deng2009imagenet}, leveraging a simple yet effective prompt expansion technique that significantly enhanced the quality of the generated content. In addition to including established models such as Stable Diffusion 1.5, we incorporated newer models from the past 12 months.
A comparison of DRAGON with existing datasets is reported in \Cref{tab:ds-comparison}. The dataset is pre-partitioned into training and test splits to facilitate its use as a benchmark, and is organized into five subsets (Extra-Small, Small, Regular, Large, Extra-Large) ranging from 2,500 to 2,500,000 training images, enabling its application across various scenarios, including few-shot learning. Beyond detailing the dataset and its generation process, this paper demonstrates its utility by presenting extensive experiments conducted on the dataset using state-of-the-art detection and attribution systems.

\begin{table}[t]
\centering
\caption{Comparison with state-of-the-art diffusion image datasets highlights the advantages of DRAGON. The proposed dataset offers significantly larger scale and higher average image quality (MPS~\cite{zhang2024learning}) with respect to the state of the art.}\label{tab:ds-comparison}
\adjustbox{width=\columnwidth}{
\begin{tabular}{l|cccccc}
\toprule
Dataset & \makecell{\# Diffusion \\ Models} & \makecell{Most Recent \\ Model} & \makecell{\# Real \\ Images} & \makecell{\# Fake \\ Images} & \makecell{Train/Test \\ split} & MPS $\uparrow$ \\
\midrule
CiFAKE \cite{bird2024cifake} & 1 & 2022 & 60 000 & 60 000 & \ding{51} & 1.55 \\
DiffusionForensics \cite{wang2023dire} & 11 & 2023 & 134 000 & 481 200 & \ding{51} & 5.93 \\
Synthbuster \cite{bammey2023synthbuster} & 9 & 2023 & 1 000 & 9 000 & \ding{55} & 9.70 \\
GenImage \cite{zhu2023genimage} & 7 & 2023 & \textbf{1 331 167} & 1 350 000 & \ding{51} & 2.83 \\ \midrule
DRAGON & \textbf{25} & \textbf{2025} & \textbf{1 331 167} & \textbf{2 600 000} & \ding{51} & \textbf{14.02}\\
\bottomrule
\end{tabular}}
\end{table}

\section{DRAGON dataset}\label{sec:dataset}

The DRAGON dataset is a large-scale synthetic image collection designed to support research in forensic tasks involving diffusion models. It comprises 2,600,000 synthetic images, representing diverse visual content generated by 25 distinct generative models, along with 1,331,167 real images sourced from ImageNet~\cite{deng2009imagenet}. Of these, 100,000 synthetic images and 50,000 real images were set aside for the test set, leaving 2,500,000 synthetic images and 1,281,167 real images available for training. DRAGON is available at \url{https://huggingface.co/datasets/lesc-unifi/dragon}.

\begin{figure}
    \centering
    \includegraphics[width=\textwidth]{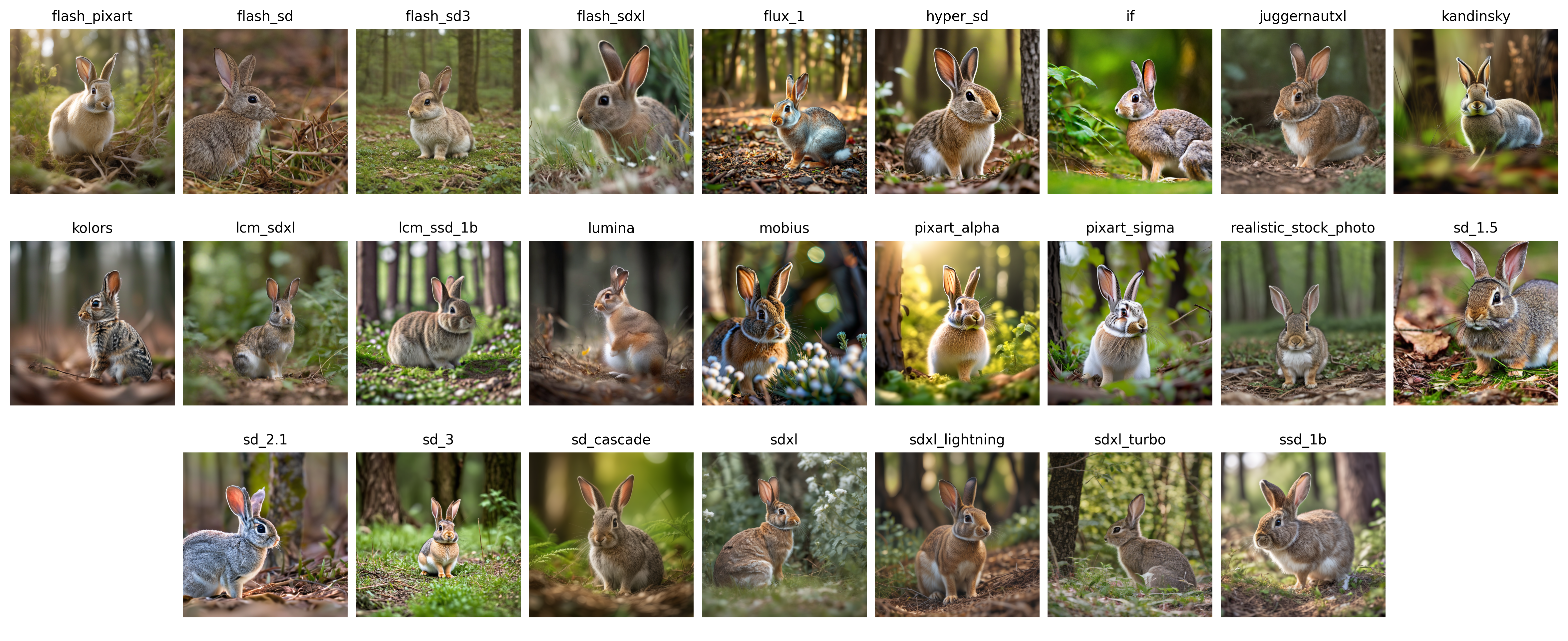}
    \caption{DRAGON dataset example for the prompt ``\textit{woodland scene featuring a cottontail rabbit in its natural habitat, soft focus background of trees and undergrowth, high resolution detailing the fur texture, close-up shot capturing the distinct white tuft on its tail, using Canon EOS R5, wide lens}'' across all models.}
    \label{fig:dataset-example}
\end{figure}

\subsection{Generative models}

To generate the images, we selected a diverse set of generative models, starting with popular latent diffusion models based on U-Net architectures, including Stable Diffusion v1.5/v2.1~\cite{rombach2022high}, Stable Diffusion XL~\cite{podell2024sdxl}, Stable Cascade~\cite{pernias2024wrstchen}, Kandinsky v3~\cite{arkhipkin2023kandinsky}, Mobius~\cite{corcel2024mobius}, and Kolors~\cite{kolors2024kolors}. We also incorporated models leveraging Diffusion Transformer architectures, such as Stable Diffusion v3~\cite{esser2024scaling}, PixArt-$\alpha$~\cite{chen2024pixartalpha}, PixArt-$\Sigma$~\cite{chen2024pixartsigma}, Lumina~\cite{gao2024lumina}, and Flux.1 schnell~\cite{blackforest2024flux}. In addition, we included DeepFloyd IF~\cite{deepfloyd2023if}, a model that operates directly in pixel space.
Moreover, we incorporated variations of the aforementioned base models, including versions distilled to operate with fewer diffusion steps — namely Stable Diffusion XL Turbo~\cite{sauer2024adversarial}, Stable Diffusion XL Lightning~\cite{lin2024sdxl}, Segmind Stable Diffusion 1B~\cite{gupta2024progressive}, and Flash variants of Stable Diffusion v1.5, Stable Diffusion XL, Stable Diffusion 3, and PixArt-$\alpha$~\cite{chadebec2025flash}. We further included models based on latent consistency techniques~\cite{luo2023latent}, such as LCM SSD-1B, LCM SDXL, and Hyper Stable Diffusion~\cite{ren2024hypersd}.
Finally, to enhance the realism of the dataset and better reflect real-world use cases, we incorporated a selection of user-finetuned models, including Juggernaut XL v8~\cite{rundiffusion2024juggernaut} and Realistic Stock Photo~\cite{yntec2024realistic}.

\subsection{Prompt generation}\label{sec:prompt-expansion}
DRAGON was developed to provide data for training tools capable of distinguishing between generated and real content. Consequently, its images must represent a broad variety of subjects to emulate real-world scenarios as closely as possible. Following the example of GenImage~\cite{zhu2023genimage}, we base our prompt construction on the 1,000 labels from ImageNet~\cite{deng2009imagenet}, each of which is used to generate 100 training images and 4 test images per model.

\begin{figure}
    \centering
    \includegraphics[width=\textwidth]{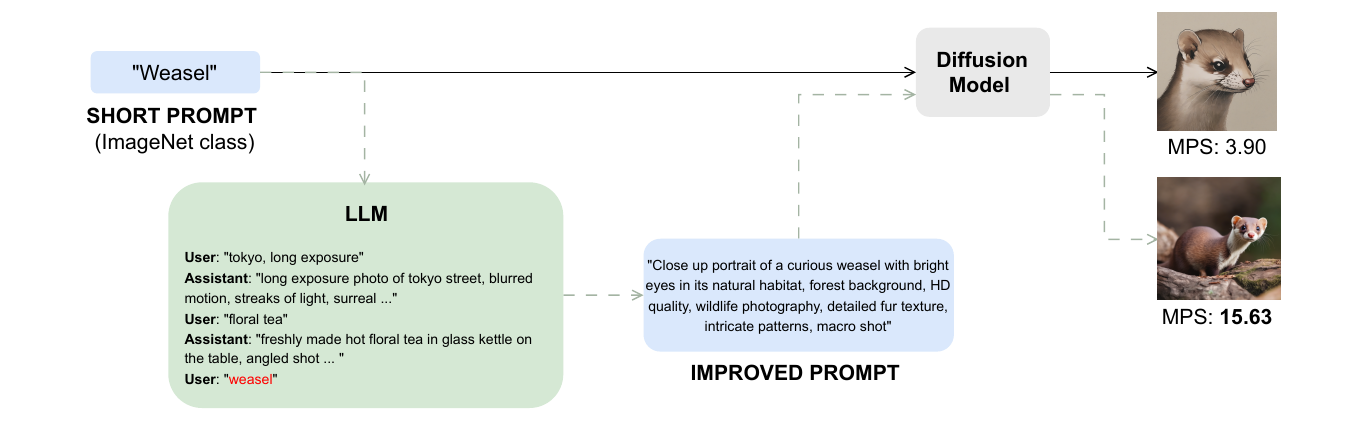}
    \caption{DRAGON prompt expansion pipeline. The black path shows the baseline approach, in which the label is used directly as the prompt. The green dashed path shows the enhanced approach, where a LLM, guided by multiple seed examples, expands the label into a higher-quality prompt. The improved prompt yields a higher Multi-dimensional Preference Scoring (MPS).}
    \label{fig:dragon-pipeline}
\end{figure}

Commonly used approaches to convert labels into prompts rely on fixed templates~\cite{bird2024cifake,zhu2023genimage,wang2023dire}, which frequently result in low-detail images that lack visual realism. Instead, we adopt a simple yet effective prompt expansion mechanism that enhances the realism of the generated outputs.
The expansion system adopted in this study is based on an few-shot in-context learning approach~\cite{brown2020language,dong2022survey}. In this framework, the large language model (LLM) is provided with several pairs of label-prompt expansions, which serve as seed examples, and is subsequently tasked with generating an expansion for a novel label.
For this task, we selected Phi-3~\cite{abdin2024phi} as the LLM and manually curated 13 prompts from those recommended on online enthusiast platforms, based on their demonstrated effectiveness in producing high-quality images.
A diagram of the expansion pipeline is reported in \Cref{fig:dragon-pipeline}.

\subsection{Subsets}

DRAGON training set is structured into five subsets (ExtraSmall, Small, Regular, Large, ExtraLarge), each containing an increasingly larger number of images to accommodate a range of research scenarios. Each subset extends the image collection of the immediately smaller subset; consequently, ExtraSmall is a strict subset of Small, which in turn is a strict subset of Regular, and so on. The largest subset, ExtraLarge, comprises 100,000 images per model, resulting in a total of 2.5 million training images, thus offering a viable solution for training large-scale models. In contrast, the smallest subset, ExtraSmall, includes 10 images per model, totaling 250 training images, making it suitable for few-shot learning settings.

The DRAGON test set, by contrast, consists of 4,000 images per model, resulting in a total of 100,000 test images. While the full test set can be used to evaluate the performance of models trained on any DRAGON subset, two additional evaluation subsets comprising 1,000 and 10,000 images are also provided. These are specifically aligned with the ExtraSmall and Small/Regular subsets, respectively. The images in these subsets are generated exclusively using the same ImageNet classes present in the corresponding training sets, thereby enabling the evaluation of performance differences between seen and unseen semantic content.

A summary of the composition of DRAGON's subsets is reported in \Cref{tab:training-size}.

\subsection{Data generation}

All images were generated using the \texttt{diffusers} library by HuggingFace~\cite{vonplaten2022diffusers}. For each model, default generation parameters (e.g., diffusion steps, resolution, etc.) were employed, based on the assumption that these defaults represent both the most commonly used settings and those that typically yield the highest image quality. For most models, the default resolution is $1024\times1024$ pixels, while Realistic Stock Photo~\cite{yntec2024realistic} and Stable Diffusion 2.1~\cite{rombach2022high} generate $768\times768$ images. The lowest available resolution is $512\times512$ pixels, provided by Stable Diffusion 1.5~\cite{rombach2022high}, Stable Diffusion XL Turbo~\cite{sauer2024adversarial}, and Flash Stable Diffusion~\cite{chadebec2025flash}.
To ensure reproducibility, each image is annotated with the model used for generation, the corresponding prompt, and the random seed.
The dataset was generated on a compute cluster equipped with NVIDIA A100 Tensor Core GPUs. On this hardware, the generation of the entire dataset required approximately 5,300 GPU hours. The average generation time per image varied significantly across models, ranging from less than one second for Flash Stable Diffusion and Stable Diffusion XL Turbo, to over 30 seconds for models such as Kolors and Lumina.

\begin{table}[tb]
\centering
\caption{Characteristics of the DRAGON subsets. Each subset in DRAGON includes all images from the immediately smaller subset, effectively extending it. This structure applies not only to the training set but also to the test set, which is provided in three subsets of increasing size.}\label{tab:training-size}
\adjustbox{width=0.7\textwidth}{
\begin{tabular}{r|ccccc}
\toprule
\multicolumn{1}{l}{}          & \textbf{XS} & \textbf{S} & \textbf{R} & \textbf{L} & \textbf{XL} \\ \midrule
\textbf{\# Prompts}           & 10          & 100        & 100        & 1,000      & 1,000       \\
\textbf{\# Training Images per Prompt} & 1           & 1          & 10         & 10         & 100         \\
\textbf{\# Training Images per Model}  & 10          & 100        & 1,000      & 10,000     & 100,000     \\
\textbf{Training Size}          & 250         & 2,500      & 25,000     & 250,000     & 2,500,000   \\ \midrule
\textbf{\# Test Images per Prompt} & 4           & 4          & 4          & 4          & 4           \\
\textbf{\# Test Images per Model}  & 40          & 400        & 400        & 4,000      & 4,000       \\
\textbf{Test size}                 & 1,000       & 10,000     & 10,000     & 100,000    & 100,000     \\ \bottomrule
\end{tabular}}
\end{table}

\section{DRAGON Analysis}\label{sec:analysis}

\subsection{Quality Evaluation}\label{sec:quality-evaluation}

In this section, we analyze the images contained in DRAGON to assess whether the prompt expansion mechanism described in \Cref{sec:prompt-expansion} yields measurable improvements in image quality. To this end, we employed state-of-the-art quality assessment metrics for text-to-image generation to compare images generated with and without prompt expansion.

\begin{figure}[tb]
    \centering
    \includegraphics[width=0.8\textwidth]{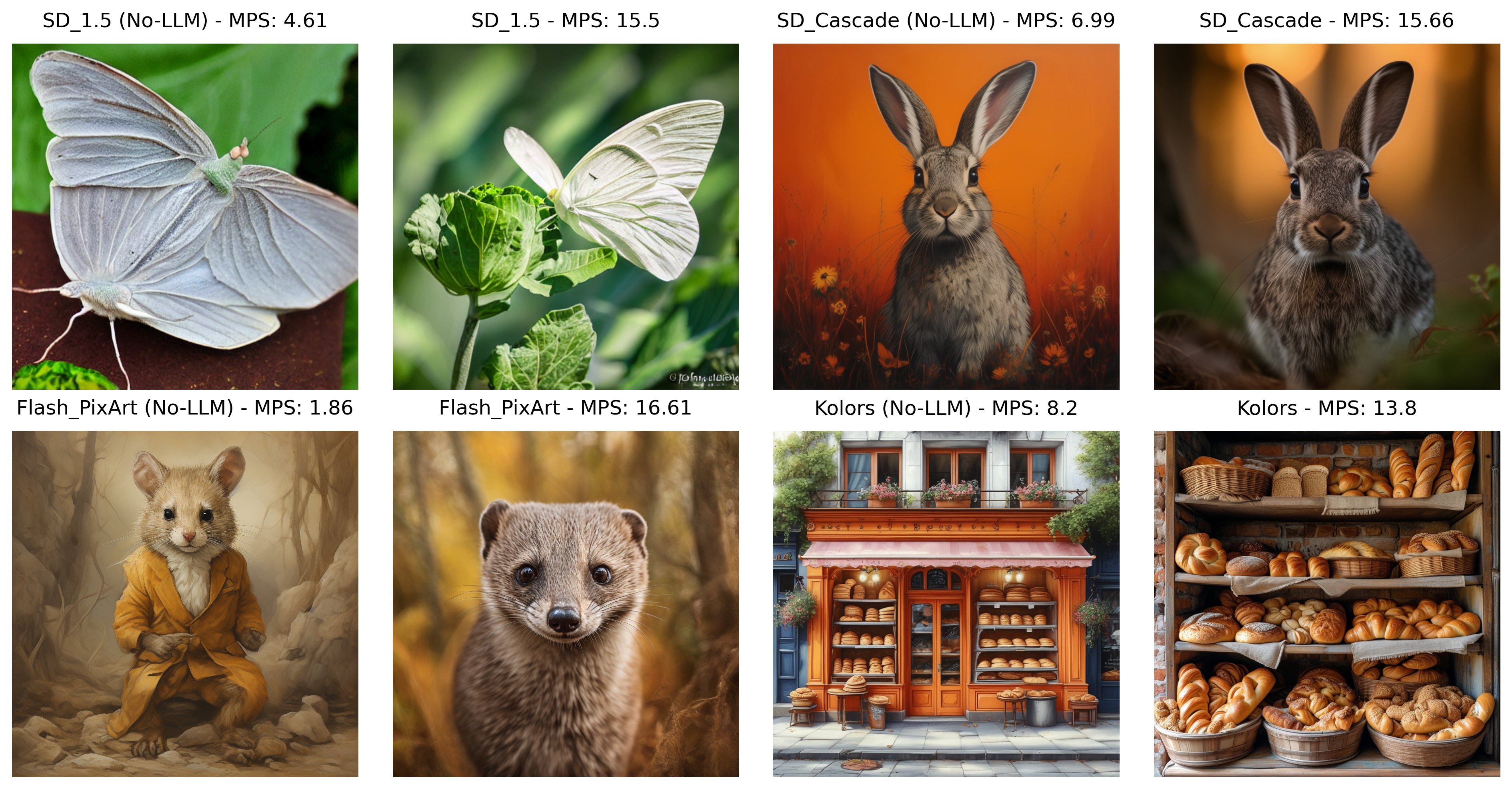}
    \caption{Comparison of image quality (MPS score) with and without LLM-based prompt expansion. For each diffusion model, the left image is generated using the original ImageNet label as the prompt, while the right image is generated using the corresponding LLM-expanded prompt.}
    \label{fig:visual-examples}
\end{figure}

Specifically, we constructed a set of images using the same generation pipeline and the same 100 ImageNet classes as in the Regular subset, but without applying the prompt expansion step, relying instead on the original class labels. For each prompt, we generated three images, resulting in 300 images per model and a total of 7,500 images. Examples of images generated with and without prompt expansion are shown in \Cref{fig:visual-examples}.

We then evaluated the quality of each image using three state-of-the-art scoring models designed to approximate human judgment in text-to-image evaluation: Human Preference Score v2 (HPS)~\cite{wu2023human}, ImageReward (IR)~\cite{xu2023imagereward}, and Multi-dimensional Preference Scoring (MPS)~\cite{zhang2024learning}. HPS and IR are single-score models trained on large-scale human preference datasets to capture overall user preferences between image pairs generated from the same prompt. In contrast, MPS offers a more fine-grained assessment by modeling human preferences across four distinct dimensions: aesthetics, semantic alignment, detail quality, and overall impression. This multidimensional approach has been shown to outperform single-score metrics in reflecting human evaluations.

As reported in \Cref{tab:quality-evaluation}, all three metrics indicate an overall improvement when prompt expansion is employed. While HPS and IR show only marginal gains, MPS reveals a substantial increase in image quality when our proposed pipeline is used. Given the superior accuracy and alignment with human judgment demonstrated by MPS compared to HPS and IR, these results suggest that our prompt expansion strategy leads to meaningful improvements in the perceptual quality of generated images.

 \begin{table}[tb]
    \centering
    \caption{Comparison of image quality in terms of SoTA image quality metrics for text-to-image generation (HPS, IR, and MPS) between the DRAGON-R test dataset and a subset of images (No-LLM) generated without using the prompt expansion technique.  }\label{tab:quality-evaluation}
    \adjustbox{width=\columnwidth}{
    \begin{tabular}{l|clclcl}
    \toprule
        \textbf{Model} & $\textbf{HPS}_{\text{\textbf{No-LLM}}} \uparrow$ & $\textbf{HPS}_{\text{\textbf{DRAGON}}} \uparrow$ & $\textbf{IR}_{\text{\textbf{No-LLM}}} \uparrow$ & $\textbf{IR}_{\text{\textbf{DRAGON}}} \uparrow$ & $\textbf{MPS}_{\text{\textbf{No-LLM}}} \uparrow$ & $\textbf{MPS}_{\text{\textbf{DRAGON}}} \uparrow$ \\ \midrule
         Stable Diffusion 1.5~\cite{rombach2022high}&  0.255 & 0.245 \textcolor{red}{(-0.010)} & -1.354 & -1.304 \textcolor{mygreen}{(+0.050)} & 4.445 & 12.423 \textcolor{mygreen}{(+7.987)} \\
         Stable Diffusion 2.1~\cite{rombach2022high}&  0.258 & 0.249 \textcolor{red}{(-0.009)} & -1.343 & -1.354 \textcolor{red}{(-0.011)} & 4.739 & 12.763 \textcolor{mygreen}{(+8.024)} \\ \midrule
        Stable Diffusion XL~\cite{podell2024sdxl} & 0.255 & 0.275 \textcolor{mygreen}{(+0.020)} & -1.280 & -1.204 \textcolor{mygreen}{(+0.076)} & 5.264 & 14.366 \textcolor{mygreen}{(+9.102)} \\
        PixArt Alpha~\cite{chen2024pixartalpha} &  0.264 & 0.285 \textcolor{mygreen}{(+0.021)} & -0.969 & -1.064 \textcolor{red}{(-0.095)} & 5.210 & 14.313 \textcolor{mygreen}{(+9.103} \\
        IF~\cite{deepfloyd2023if} &  0.256 & 0.261 \textcolor{mygreen}{(+0.005)} & -1.298 & -1.307 \textcolor{red}{(-0.009)} & 4.754 & 13.435 \textcolor{mygreen}{(+8.681)} \\
        Kandinsky 3~\cite{arkhipkin2023kandinsky} &  0.262 & 0.274 \textcolor{mygreen}{(+0.012)} & -1.093 & -1.103 \textcolor{red}{(-0.010)} & 5.689 & 14.408 \textcolor{mygreen}{(+8.719)} \\ \midrule
        Stable Diffusion 3~\cite{esser2024scaling} & 0.272 & 0.282 \textcolor{mygreen}{(+0.010)} & -1.284 & -1.240 \textcolor{mygreen}{(+0.044)} & 5.780 & 14.478 \textcolor{mygreen}{(+8.698)} \\
        Stable Diffusion XL Turbo~\cite{sauer2024adversarial}  & 0.268 & 0.271 \textcolor{mygreen}{(+0.003)} & -1.230 & -1.194 \textcolor{mygreen}{(+0.036)} & 5.689 & 14.346 \textcolor{mygreen}{(+8.701)} \\
        Stable Diffusion XL Lightning~\cite{lin2024sdxl} & 0.268 & 0.283 \textcolor{mygreen}{(+0.015)} & -1.204 & -1.142 \textcolor{mygreen}{(+0.062)} & 5.900 & 14.565 \textcolor{mygreen}{(+8.665)} \\
         Stable Diffusion Cascade~\cite{pernias2024wrstchen} & 0.274 & 0.281 \textcolor{mygreen}{(+0.007)} & -1.030 & -1.087 \textcolor{red}{(-0.057)} & 6.634 & 14.876 \textcolor{mygreen}{(+8.242)} \\
         Hyper Stable Diffusion~\cite{ren2024hypersd} & 0.295 & 0.307 \textcolor{mygreen}{(+0.012)} & -0.886 & -0.885 \textcolor{mygreen}{(+0.001)} & 6.262 & 14.821 \textcolor{mygreen}{(+8.559)} \\
        PixArt Sigma~\cite{chen2024pixartsigma} &  0.277 & 0.289 \textcolor{mygreen}{(+0.021)} & -1.065 & -1.143 \textcolor{red}{(-0.078)} & 5.299 & 14.379 \textcolor{mygreen}{(+9.080)} \\
        Segmind Stable Diffusion 1B~\cite{gupta2024progressive} &  0.252 & 0.280 \textcolor{mygreen}{(+0.028)} & -1.295 & -1.138 \textcolor{mygreen}{(+0.157)} & 5.141 & 14.274 \textcolor{mygreen}{(+9.133)} \\
        Latent Consistency Model SSD-1B~\cite{luo2023latent} & 0.249 & 0.260 \textcolor{mygreen}{(+0.011)} & -1.264 & -1.145 \textcolor{mygreen}{(+0.119)} & 4.479 & 13.226 \textcolor{mygreen}{(+8.747)} \\
        Latent Consistency Model SDXL~\cite{luo2023latent} & 0.255 & 0.261 \textcolor{mygreen}{(+0.006)} & -1.264 & -1.229 \textcolor{mygreen}{(+0.035)} & 5.006 & 13.895 \textcolor{mygreen}{(+8.979)} \\
        JuggernautXL v8~\cite{rundiffusion2024juggernaut} & 0.266 & 0.282 \textcolor{mygreen}{(+0.016)} & -1.293 & -1.168 \textcolor{mygreen}{(+0.125)} & 5.815 & 14.752 \textcolor{mygreen}{(+8.937)} \\
        Realistic Stock Photo~\cite{yntec2024realistic} & 0.259 & 0.257 \textcolor{red}{(-0.002)} & -1.316 & -1.276 \textcolor{mygreen}{(+0.040)} & 4.625 & 12.882 \textcolor{mygreen}{(+8.257)} \\
         Mobius~\cite{corcel2024mobius} & 0.285 & 0.307 \textcolor{mygreen}{(+0.022)} & -1.094 & -1.077 \textcolor{mygreen}{(+0.017)} & 5.866 & 14.749 \textcolor{mygreen}{(+8.883)} \\
        Lumina~\cite{gao2024lumina} & 0.245 & 0.262 \textcolor{mygreen}{(+0.017)} & -1.037 & -1.159 \textcolor{red}{(-0.122)} & 4.099 & 13.293 \textcolor{mygreen}{(+9.194)} \\
        Flux.1 schnell~\cite{blackforest2024flux} & 0.273 & 0.288 \textcolor{mygreen}{(+0.015)} & -1.248 & -1.151 \textcolor{mygreen}{(+0.097)} & 6.036 & 14.737 \textcolor{mygreen}{(+8.701)}\\
         Kolors~\cite{kolors2024kolors} & 0.277 & 0.288 \textcolor{red}{(+0.011)} & -0.942 & -1.032 \textcolor{red}{(-0.090)} & 5.849 & 14.426 \textcolor{mygreen}{(+8.577)} \\
         Flash Stable Diffusion~\cite{chadebec2025flash} & 0.239 & 0.232 \textcolor{red}{(-0.007)} & -1.414 & -1.305 \textcolor{mygreen}{(+0.109)} & 4.035 & 12.241 \textcolor{mygreen}{(+8.206)} \\
        Flash Stable Diffusion XL~\cite{chadebec2025flash} & 0.256 & 0.260 \textcolor{mygreen}{(+0.004)} & -1.288 & -1.236 \textcolor{mygreen}{(+0.052)} & 5.315 & 13.727 \textcolor{mygreen}{(+8.412)} \\
        Flash Stable Diffusion 3~\cite{chadebec2025flash} & 0.257 & 0.260 \textcolor{mygreen}{(+0.003)} & -1.286 & -1.202 \textcolor{mygreen}{(+0.084)} & 5.395 & 14.199 \textcolor{mygreen}{(+8.804)} \\
        Flash PixArt~\cite{chadebec2025flash}  & 0.238 & 0.262 \textcolor{mygreen}{(+0.024)} & -1.003 & -1.090 \textcolor{red}{(-0.095)} & 4.141 & 13.616 \textcolor{mygreen}{(+9.475)} \\
        \bottomrule
    \end{tabular}}
\end{table}

\subsection{Forensic Analysis}

The primary purpose of the proposed dataset is to support researchers in developing methods for analyzing potentially generated images in order to understand their lifecycle, which emerged as a significant challenge in the digital forensics community in recent years~\cite{tariang2024synthetic,lin2024detecting}.
In the following subsections, we first conduct a frequency analysis to qualitatively assess the traces left by the generative models. We then evaluate the performance of state-of-the-art methods on the DRAGON dataset, focusing on two core tasks: synthetic image detection, which involves distinguishing real images from generated ones; and model attribution, which seeks to identify the specific generative model responsible for producing a given image.

\subsubsection{Frequency Analysis}\label{sec:freq-analysis}

\begin{figure}
    \centering
    \includegraphics[width=\textwidth]{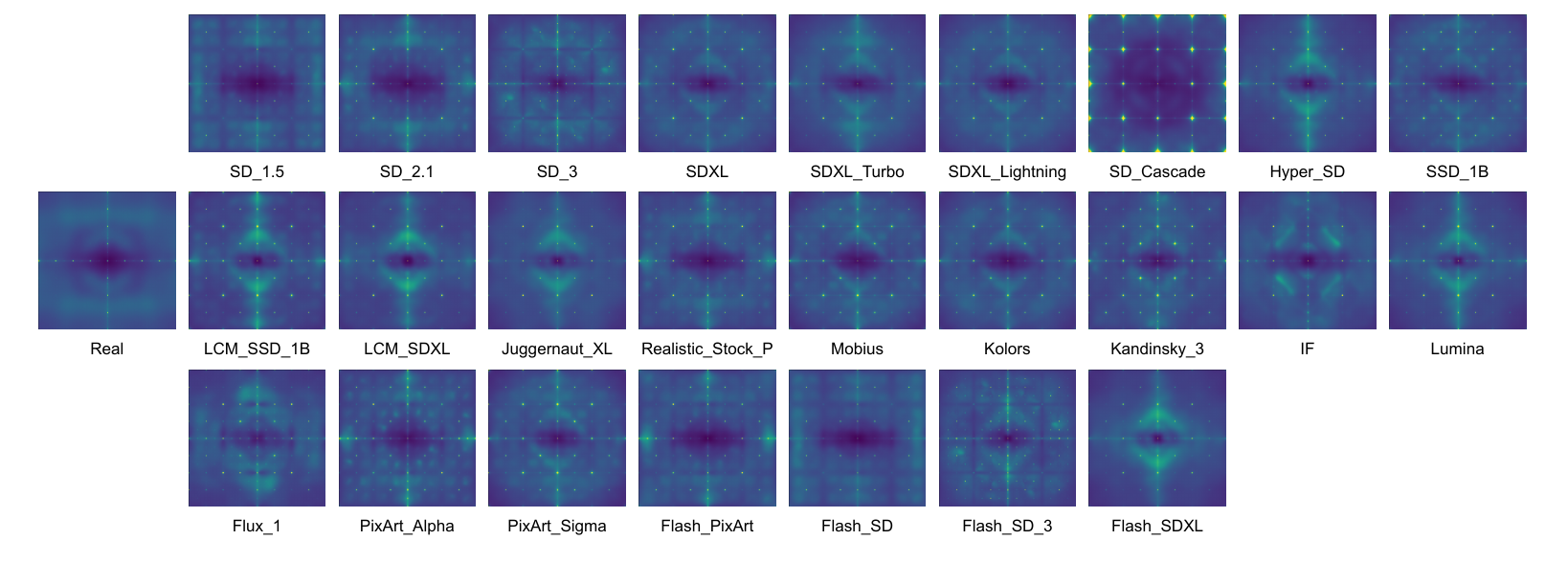}
    \caption{Fourier transform (amplitude) of the average of 1000 noise residuals for each model.}
    \label{fig:fourier_transform}
\end{figure}

Recent studies suggest that AI-generated media often retain distinctive frequency-domain signatures that can be exploited for attribution tasks~\cite{corvi2023intriguing,wang2020cnn}. In \Cref{fig:fourier_transform}, we present the average Fourier transform amplitude for each model in the DRAGON dataset, computed over 1,000 noise residuals, following the methodology proposed by \citet{corvi2023intriguing}. The spectrum labeled \textit{Real} corresponds to a random sample of 1,000 images drawn from the ImageNet validation set.

This qualitative analysis reveals that generative models produce spectral patterns that are rarely observed in real images. Moreover, substantial differences are evident between the spectra of images generated by different models. While such discrepancies are expected between unrelated architectures (e.g., Stable Diffusion XL vs. Flux.1), we also observe notable distinctions between base models and their distilled variants (e.g., PixArt-$\alpha$ vs. Flash PixArt-$\alpha$), as well as between base models and their fine-tuned counterparts (e.g., Stable Diffusion XL vs. Juggernaut\_XL).

From a forensic standpoint, these subtle yet consistent spectral differences are particularly valuable, as they may facilitate the identification of specific generative models, including those potentially used to produce targeted misinformation or propaganda.

\subsubsection{Synthetic image detection}

In the following analysis, we assess the detection performance of several well-established forensic methods on the Regular test subset of DRAGON (DRAGON-R).
DE-FAKE~\cite{sha2023fake} detects and attributes fake images by leveraging the joint image-text embeddings of CLIP~\cite{radford2021learning}, using a binary classifier for detection and a multi-class one for attribution. DIRE~\cite{wang2023dire} identifies diffusion-generated images by measuring the reconstruction error when inverting and then denoising an image, exploiting the tendency of diffusion models to better reconstruct their own outputs. CLIPDet~\cite{cozzolino2024raising} relies on a small set of paired real and fake images, using CLIP-ViT features combined with a linear SVM for detection. UnivFD~\cite{ojha2023towards} performs detection using the fixed feature space of a pre-trained CLIP-ViT model, employing either nearest neighbor classification or linear probing to identify synthetic content.

\paragraph{Baseline pretrained models}

In this initial experiment, we evaluated the performance of the four selected models using the pretrained versions provided by their respective authors. These models were originally trained on a smaller and older set of diffusion models compared to those featured in DRAGON.
Real images from ImageNet are JPEG-compressed, while DRAGON's synthetic images are in lossless PNG format -- a difference that may introduce compression-related biases~\cite{grommelt2024fake}. To address this, we evaluated performance on both the original PNG images and JPEG-compressed versions (quality factor 96, matching the average quality in ImageNet). Additionally, to investigate whether increased image realism affects the discriminative power of forensic detectors, we evaluated each method on a variant of the dataset introduced in \Cref{sec:quality-evaluation}, in which the prompt expansion step described in \Cref{sec:prompt-expansion} was omitted.

The evaluation results are presented in \Cref{tab:zero-shot}. While DE-FAKE and CLIPDet show minimal sensitivity to JPEG compression, UnivFD exhibits a slight degradation in performance, and DIRE experiences a significant drop in accuracy when tested on JPEG-compressed synthetic images. This suggests that DIRE’s strong performance may partially depend on compression artifacts rather than genuine generation traces.
Moreover, most of the evaluated forensic methods exhibit a modest performance drop when comparing the No-LLM and DRAGON-R subsets using PNG images. In contrast, performance remains largely consistent between these subsets when using JPEG-compressed images. These results indicate that, despite the substantial increase in visual realism in DRAGON images compared to No-LLM images (as measured by MPS), the forensic methods under study are only marginally sensitive to these realism improvements.

\begin{table}[t]
\centering
\caption{Performance of pre-trained synthetic image detection methods on the DRAGON-R subset, and on a corresponding set of images generated without LLM-based prompt expansion. All experiments were repeated using JPEG-compressed synthetic images with a quality factor of 96.}
\label{tab:zero-shot}
\adjustbox{width=0.7\columnwidth}{
\begin{tabular}{l|cc|cc}
\toprule
\textbf{Detector} & \textbf{No-LLM} & \textbf{DRAGON-R} & $\textbf{No-LLM}_{\text{\textbf{JPG}}}$ & $\textbf{DRAGON-R}_{\text{\textbf{JPG}}}$\\ \midrule
DE-FAKE \cite{sha2023fake}& 0.86 & 0.87 & 0.86 & 0.87 \\
DIRE \cite{wang2023dire} & 0.99 & 0.92 & 0.83 & 0.83 \\
CLIPDet \cite{cozzolino2024raising} & 0.77 & 0.75 & 0.77 & 0.76\\
UnivFD \cite{ojha2023towards} & 0.63 & 0.61 & 0.59 & 0.57 \\
\bottomrule
\end{tabular}}
\end{table}

\paragraph{Impact of training on DRAGON}

In this experiment, we retrained selected detection methods on the DRAGON dataset to assess the impact on performance. Among the four available methods, only DE-FAKE and UnivFD could be retrained. CLIPDet was excluded from this analysis due to the unavailability of training code from the original authors. DIRE was also omitted, as its computational requirements exceeded our available resources, requiring over 10 GPU-hours for inference alone, rendering full retraining infeasible within our time constraints.

Moreover, we evaluated the robustness of the detection methods under more realistic and challenging conditions, simulating degradations typically introduced during image transmission and storage. To this end, we analyzed the performance of both pretrained and retrained detectors on images from the DRAGON-R subset after applying JPEG compression with quality factors ranging from 90 to 10, as well as after resizing the images to resolutions of $256 \times 256$ and $128 \times 128$ using Lanczos resampling. The results of this robustness evaluation are presented in \Cref{tab:zero-shot-robust}.

In the JPEG compression analysis, DE-FAKE was the most resilient, maintaining a balanced accuracy of 0.86 even at the lowest quality. CLIPDet showed a notable drop of about 0.12 at quality factor 30, while UnivFD performed near chance across most levels. Similar patterns emerged in the resizing tests: DE-FAKE remained stable, UnivFD stayed near chance, DIRE showed a 0.20 accuracy decrease, and CLIPDet struggled the most, dropping to 0.63 accuracy at $128 \times 128$ resolution.

Retraining the models on the DRAGON dataset led to significant performance improvements, particularly for DE-FAKE and UnivFD, which achieved gains of 0.10 and 0.20 in balanced accuracy, respectively. These improvements were consistent across all robustness scenarios.
These results underscore the importance of training detection models on up-to-date datasets that reflect the characteristics of modern generative models.

\begin{table}[t]
\centering
\caption{Robustness analysis of pre-trained detection methods on DRAGON-R.  * denotes re-trained detectors with the official codes. We compare performance on original (undegraded) images with results obtained after JPEG compression at varying quality factors, as well as after image resizing.}
\label{tab:zero-shot-robust}
\adjustbox{width=0.7\columnwidth}{
\begin{tabular}{lc|ccccc|cc}
\toprule
\multirow{2}{*}{\textbf{Detector}} &  & \multicolumn{5}{c|}{\textbf{JPEG compression (QF)}} & \multicolumn{2}{c}{\textbf{Resize}} \\
 & Baseline & 90 & 70 & 50 & 30 & 10 & 256 & 128 \\ \midrule
\textbf{DE-FAKE} \cite{sha2023fake} & 0.868 & 0.870 & 0.868 & 0.868 & 0.869 & 0.869 & 0.867 & 0.863 \\
\textbf{DIRE} \cite{wang2023dire}& 0.916 & 0.899 & 0.979 & 0.868 & 0.745 & 0.778 & 0.716 & 0.740 \\
\textbf{CLIPDet} \cite{cozzolino2024raising} & 0.753 & 0.788 & 0.766 & 0.772 & 0.658 & 0.620 & 0.855 & 0.638 \\
\textbf{UnivFD} \cite{ojha2023towards} & 0.608 & 0.558 & 0.561 & 0.551 & 0.552 & 0.546 & 0.511 & 0.518 \\
\midrule
$\textbf{DE-FAKE}^*$ \cite{sha2023fake} & 0.986 & 0.985 & 0.984 & 0.985 & 0.984 & 0.984 & 0.985 & 0.984 \\
$\textbf{UnivFD}^*$ \cite{ojha2023towards} & 0.807 & 0.806 & 0.856 & 0.837 & 0.818 & 0.772 & 0.781 & 0.781 \\
\bottomrule
\end{tabular}}
\end{table}

\subsubsection{Model attribution}

In this final experiment, we evaluated model attribution performance. Among the selected state-of-the-art methods, only DE-FAKE supports the attribution of images to their source generative model. Consequently, this analysis was conducted exclusively using DE-FAKE, trained on DRAGON-R.

\begin{figure}
    \centering
    \includegraphics[width=\textwidth]{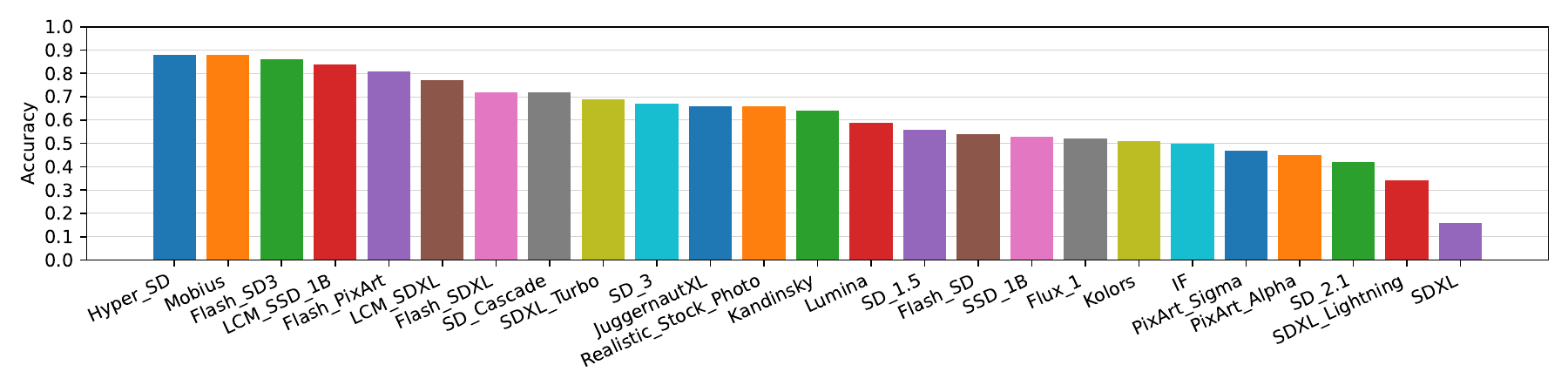}
    \caption{Model-wise attribution accuracy of DE-FAKE trained on the DRAGON-R training set. The average accuracy across all models is 0.62.}
    \label{fig:attribution}
\end{figure}

The detector achieved an average classification accuracy of 0.62 across the 25 generative models, demonstrating reasonable effectiveness in distinguishing between different generators. Model-wise attribution accuracies are presented in \Cref{fig:attribution}, with models ordered by decreasing performance. Accuracy varied substantially across models, ranging from 0.88 to 0.16, indicating that some models are considerably easier to identify than others.

Stable Diffusion XL exhibited the lowest attribution accuracy, making it the most challenging model to distinguish. Interestingly, it is frequently misclassified as one of its distilled variants (e.g., SDXL Turbo, SDXL Lightning) or fine-tuned derivatives (e.g., JuggernautXL v8, Realistic Stock Photo). This suggests that while these models share a common base architecture, the distillation and fine-tuning processes introduce distinguishable artifacts or ``fingerprints'' that enhance attribution performance for the derived versions.

More broadly, many attribution errors occur among closely related models—for instance, PixArt-$\alpha$ and PixArt-$\Sigma$ are often confused with one another. Nonetheless, as discussed in \Cref{sec:freq-analysis}, our qualitative frequency-domain analysis reveals consistent spectral differences even between such closely related models. These findings highlight the potential for further improving attribution methods to better capture subtle, model-specific cues.

\section{Limitations}\label{sec:limitations}

Although we have made significant efforts to create a dataset that is as useful as possible for the scientific community a number of limitations remain in our work.

\paragraph{Generative models}
The dataset includes images generated by 25 models, most of which were released within the past year. We aimed to balance well-known, widely adopted models with lesser-known alternatives. However, due to the network effect associated with the popularity of certain models, there is an overrepresentation of variants derived from a common base architecture. For instance, Stable Diffusion XL, which is included not only in its base version but also in six of its derivatives. Although these variants exhibit distinct behaviors, as shown in \Cref{sec:freq-analysis}, the images they generate may share underlying characteristics.
Moreover, the proposed dataset only contains images generated using open-weight diffusion models, and does not include any images produced by proprietary models.
Researchers using the DRAGON dataset should be aware of this potential bias.

\paragraph{ImageNet labels and prompts}
ImageNet labels are commonly used as a foundation for generating datasets with diverse content. However, like all human-annotated datasets, ImageNet and its labels are not without flaws~\cite{kisel2025flaws}, such as overlapping classes and mislabelled images. Moreover, the automated prompt expansion procedure employed in our pipeline is susceptible to misunderstandings and hallucinations, which can result in prompts that lead to content semantically different from the original label. As a result, the synthetic images generated for this dataset may not faithfully represent the ImageNet concepts used as prompts. Consequently, this dataset should not be used for content classification tasks where ImageNet classes are treated as ground-truth targets.

\paragraph{Quality of the generate images}
While the average image quality in DRAGON is significantly higher than that of currently available datasets, it is important to note that the prompt expansion mechanism does not entirely eliminate the issue of trivially identifiable generated images. A substantial number of images in DRAGON remain unrealistic; in particular, depictions of human figures often contain anatomically inconsistent elements. Consequently, the proposed dataset should not be used under the assumption that all its contents are indistinguishable from real-world imagery.

\section{Conclusion}

In this paper, we introduced DRAGON, a large-scale dataset of synthetic images generated using diffusion models. The dataset was created using 25 generative models, the majority of which were released within the past twelve months. Compared to existing state-of-the-art datasets, DRAGON includes a greater number of images and a broader variety of models, making it a valuable resource for the development of advanced systems for synthetic image detection and model attribution. Moreover, through the use of a simple prompt expansion mechanism, the generated images exhibit higher quality scores, surpassing those found in existing datasets. We evaluated the spectral signatures of generated images, revealing distinct traces for each model. Furthermore, we compared the performance of existing detection and attribution methods in both pretrained and retrained settings on DRAGON. Results demonstrate that retraining on our dataset significantly improves performance, emphasizing the value of up-to-date data.

\section*{Acknowledgments}

This work was partially supported by JSPS KAKENHI Grant JP24H00732, by JST CREST Grant JPMJCR20D3, by JST AIP Acceleration Grant JPMJCR24U3, and by JST K Program Grant JPMJKP24C2 Japan.

The authors would like to thank Cosima Raphaella Körfer, Chiara Albisani, Francesca Nizzi, and Luca Capannesi for their valuable contributions in the preparation of the dataset.

\bibliographystyle{plainnat}
\bibliography{biblio}

\end{document}